\documentclass[12pt]{article}
\usepackage[utf8]{inputenc}
\usepackage[T1]{fontenc}
\usepackage{microtype}
\usepackage{amsmath,amssymb}
\usepackage{array}
\usepackage{xcolor}
\usepackage{graphicx}
\usepackage[english]{babel}
\usepackage{csquotes}
\usepackage{hyperref}
\usepackage{booktabs}
\usepackage{longtable}
\usepackage{listings}
\usepackage{float}
\usepackage{caption}
\usepackage{titlesec}
\usepackage{setspace}
\usepackage{fancyhdr}
\usepackage{geometry}
\usepackage{xurl}
\usepackage{tabularx}
\usepackage{tcolorbox}
\usepackage{threeparttable}
\usepackage{changepage}
\usepackage{makecell}
\usepackage{enumitem}
\usepackage[utf8]{inputenc}
\DeclareUnicodeCharacter{8457}{$^\circ$F}
\DeclareUnicodeCharacter{8722}{-}
\DeclareUnicodeCharacter{8594}{$\rightarrow$}
\DeclareUnicodeCharacter{8734}{$\infty$}
\DeclareUnicodeCharacter{9733}{$\bigstar$}
\DeclareUnicodeCharacter{03B1}{$\alpha$}
\DeclareUnicodeCharacter{2264}{$\leq$}
\DeclareUnicodeCharacter{03C1}{$\rho$}

\titleformat{\section}{\normalfont\Large\bfseries}{\thesection}{1em}{}
\titleformat{\subsection}{\normalfont\large\bfseries}{\thesubsection}{1em}{}

\title{\Large Subjective Evaluation Profile Analysis of Science Fiction\\Short Stories and Its Critical-Theoretical Significance}
\author{
  \begin{tabular}{c}
    \large Kazuyoshi Otsuka \\[-0.2em]
    \normalsize Independent Researcher \\[-0.3em]
    \normalsize \texttt{otsuka@ocoinc.jp}
  \end{tabular}
}

\begin{document}
\date{}
\maketitle

\begin{abstract}
This study positions large language models (LLMs) as "subjective literary critics" to explore aesthetic preferences and evaluation patterns in literary assessment. Ten Japanese science fiction short stories were translated into English and evaluated by six state-of-the-art LLMs across seven independent sessions. Principal component analysis and clustering techniques revealed significant variations in evaluation consistency (α ranging from 1.00 to 0.35) and five distinct evaluation patterns. Additionally, evaluation variance across stories differed by up to 4.5-fold, with TF-IDF analysis confirming distinctive evaluation vocabularies for each model. Our seven-session within-day protocol using an original Science Fiction corpus strategically minimizes external biases, allowing us to observe implicit value systems shaped by RLHF and their influence on literary judgment. These findings suggest that LLMs may possess individual evaluation characteristics similar to human critical schools, rather than functioning as neutral benchmarkers.
\end{abstract}


\textbf{Data Availability:} Data can be found here: \url{https://doi.org/10.5281/zenodo.15556522}

\section{Introduction}
Large language models (LLMs) have demonstrated remarkable capabilities across various natural language tasks. However, traditional evaluations have primarily focused on objective tasks, leaving the subjective and aesthetic judgment aspects largely unexplored. While existing benchmarks (Yu et al. 2024 \cite{Yu2024}; H.Chen et al. 2025\cite{Chen2025}) have contributed to evaluating objective question-answering and creative generation, they lack the perspective of "how LLMs evaluate literary works as readers."

This research gap is particularly important in literary criticism, a domain that demands both narrative structure comprehension and aesthetic judgment capabilities—abilities traditionally considered uniquely human. In this study, we define "evaluation profile" as the totality of consistent patterns, preferences, and approaches that LLMs exhibit when reading and evaluating literary works, positioning multiple LLMs as parallel critics to explore their characteristics.

The central question of this research is: ``Do different LLMs exhibit consistent evaluation profiles in literary assessment?'' To explore this primary question, we established three sub-questions:
\begin{enumerate}
    \item Do evaluation profiles of different models show statistically distinguishable structures?
    \item How much do evaluation variances between models differ across works?
    \item What correspondences exist between evaluation language and evaluation score patterns?
\end{enumerate}

Through controlled experiments using ten science fiction short stories, we demonstrated that LLMs exhibit statistically distinguishable "evaluation profiles." These findings challenge the conventional assumption of uniform AI evaluation and suggest the possibility that multiple LLMs can provide complementary perspectives. Our methodological design addressed several critical research challenges: (i) session-timing trade-off through a same-day evaluation protocol, (ii) original-vs-existing corpus considerations by using newly created SF stories, and (iii) prompt design choices to minimize directive bias. These methodological trade-offs are further analyzed in Appendix~\ref{appendix_F}.

\section{Related Work}

\subsection{Computational Literary Analysis and Evaluation Frameworks}
The field of digital humanities has a rich tradition of computational text analysis, from Moretti (2013)\cite{Moretti2013}'s "distant reading" approach to Jockers (2013)\cite{Jockers2013}' macroanalytic methods. These foundational studies established quantitative approaches to literature that focused primarily on descriptive analysis.

Recent developments have shifted from descriptive analysis toward evaluative frameworks. Yu et al. (2024)\cite{Yu2024} introduced LFED, the first purpose-built benchmark for long-form fiction understanding with 1,304 question-answer items across 95 complete novels. H.Chen et al. (2025)\cite{Chen2025} proposed WritingBench, covering six writing domains using criterion-aware scoring. However, these studies focus on objective comprehension, differing from our subjective aesthetic evaluation approach.

\subsection{AI and Aesthetic Judgment}
Research on AI aesthetic judgment has expanded from visual arts Elgammal et al. (2017)\cite{Elgammal2017} to textual domains. In poetry evaluation, L.Chen et al. (2024)\cite{Chen2024} provided a diversity-oriented benchmark for automatic poetry generation. Mikros (2025)\cite{Mikros2025} demonstrated GPT-4o's capabilities and limitations in literary style imitation, revealing models' strengths in style mimicry.

Liu et al. (2023)\cite{Liu2023} conducted pioneering research using GPT-4 itself as an evaluator through chain-of-thought prompting. Our study extends this approach by utilizing multiple models as simultaneous critics.

\subsection{Reader-Response Theory and AI}
Reader-response theory (Rosenblatt 1978\cite{Rosenblatt1978}; Iser 1978\cite{Iser1978}) argues that meaning emerges through interaction between text and reader. While most AI literary research focuses on generation (the "writer's" perspective), our study explores LLMs as readers.
Bode (2023)\cite{Bode2023} criticizes purely computational approaches to literature, advocating for acceptance of models' subjectivity. Our approach documents the distribution of subjective responses across models rather than seeking a single "correct" interpretation.

\section{Methodology}

\subsection{Experimental Materials}
This study utilized ten Japanese science fiction short stories. This material selection was based on intentions to control multiple confounding variables.

First, the choice of science fiction genre minimized cultural bias impacts. Since stories centered on family relationships or romance strongly reflect national, cultural, and historical perspectives, we used science fiction dealing with future settings and fictional technologies to reduce direct associations with specific societal values.

Second, by writing originals in Japanese and pre-translating them, we eliminated the influence of translation ability differences between models on evaluation results. By translating all works into English through unified translation by Claude Sonnet 3.7, we separated each model's literary evaluation ability from translation processing capability, enabling pure evaluation characteristic measurement.

These works were original pieces co-written by humans under AI assistance from ChatGPT-4o and Claude Sonnet 3.7, supervised by humans. We confirmed that none existed in any language model's training data at the time of research implementation (May 2025). All works were translated from Japanese to English using Sonnet 3.7 in the same session with identical prompts.

To avoid self-evaluation bias, these two models involved in creation and translation were excluded from evaluation targets.

\subsection{Target Models for Evaluation}
We selected the following six state-of-the-art large language models as evaluation targets:
\begin{itemize}
    \item ChatGPT-4.5
    \item Gemini-2.5-flash-preview-05-20 (hereafter, Gemini-2.5)
    \item Notebook LM
    \item OpenAI-o3
    \item Grok3
    \item Claude Sonnet 4 (hereafter, Sonnet4)
\end{itemize}

Selection criteria included currency as of May 2025, diversity across different development companies, and non-involvement in translation/creation processes.

\subsection{Evaluation Protocol}
Each model was provided with a standardized prompt requesting story evaluation based on four criteria (See Appendix~\ref{appendix_B} for the complete prompt):
\begin{enumerate}
    \item Narrative structure and coherence
    \item Originality of concept or theme
    \item Emotional and aesthetic resonance
    \item Overall literary quality
\end{enumerate}

The prompt deliberately contained only the minimal role descriptor "Act as a literary critic" and the four criteria above. By withholding any further critical perspective or weighting, we sought to reveal each model's implicit value system that has emerged through pre‑training and reinforcement learning with human feedback (RLHF), rather than measuring mere obedience to detailed instructions.

Models produced a score (0-100) and a brief comment for each story. To enable precise comparative analysis, ties were disallowed and at least a 10‑point span between the highest‑ and lowest‑rated story was mandated, forcing each model to emit a complete ranking of the ten works.

Seven independent evaluation sessions were conducted for every model, all within a single calendar day. This concentrated schedule reduces long‑term drift (i.e., parameter or policy changes that accrue over days or weeks) while providing enough repetitions to estimate stability. Nevertheless, short‑interval repetition can induce evaluation fatigue (i.e., response rigidification after repeated exposure) and may magnify transient platform‑side phenomena such as cache reuse or adaptive rate‑limit windows. 

Session‑separation features supplied by each platform were used to purge chat memory between runs, and anonymized story identifiers prevented semantic priming.

\subsection{Analytical Methods}
\begin{itemize}
    \item \textbf{Principal Component Analysis (PCA)}: Applied dimensionality reduction to standardized evaluation data (42 evaluations × 10 works) to identify latent structures in evaluation profiles.
    \item \textbf{Clustering Analysis}: Applied K-means and hierarchical clustering methods, determining optimal cluster numbers through elbow method, silhouette score analysis, and dendrograms.
    \item \textbf{Evaluation Consistency Analysis}: Quantified inter-session consistency for each model using Cronbach's alpha coefficient and Spearman's rank correlation coefficient. All pairwise comparison p-values were adjusted using Benjamini-Hochberg method to FDR = 0.05 (q < 0.05 considered significant), controlling Type I error from multiple comparisons. Used pairwise t-tests for paired samples (between evaluation sessions). Applied within ranges where normality wasn't rejected by Shapiro-Wilk test; when normality assumptions were problematic, applied Wilcoxon signed-rank test. (Details in Appendix ~\ref{appendix_D3})
    \item \textbf{Text Mining}: Conducted TF-IDF analysis on evaluation comments to identify model-specific evaluation language patterns. Preprocessing excluded 50 stop words (detailed list in Appendix~\ref{appendix_E1}. Additionally conducted supplementary evaluation profile vocabulary feature analysis based on clustering results.
\end{itemize}

\section{Results}
\subsection{Hierarchical Structure of Evaluation Consistency}
We assessed the internal consistency of model evaluations by calculating Cronbach's $\alpha$ coefficients for each LLM. As summarized below and detailed in Appendix~\ref{appendix_A2}, these values revealed a clear hierarchy:

\begin{itemize}
    \item \textbf{Notebook LM}: $\alpha$=1.00 (perfect match)
    \item \textbf{Grok3}: $\alpha$=0.96 (extremely high)
    \item \textbf{Sonnet4}: $\alpha$=0.89 (high)
    \item \textbf{OpenAI-o3}: $\alpha$=0.81 (moderately high)
    \item \textbf{ChatGPT-4.5}: $\alpha$=0.69 (moderate)
    \item \textbf{Gemini-2.5}: $\alpha$=0.35 (low)
\end{itemize}

\begin{figure}[H]
    \includegraphics[width=\linewidth]{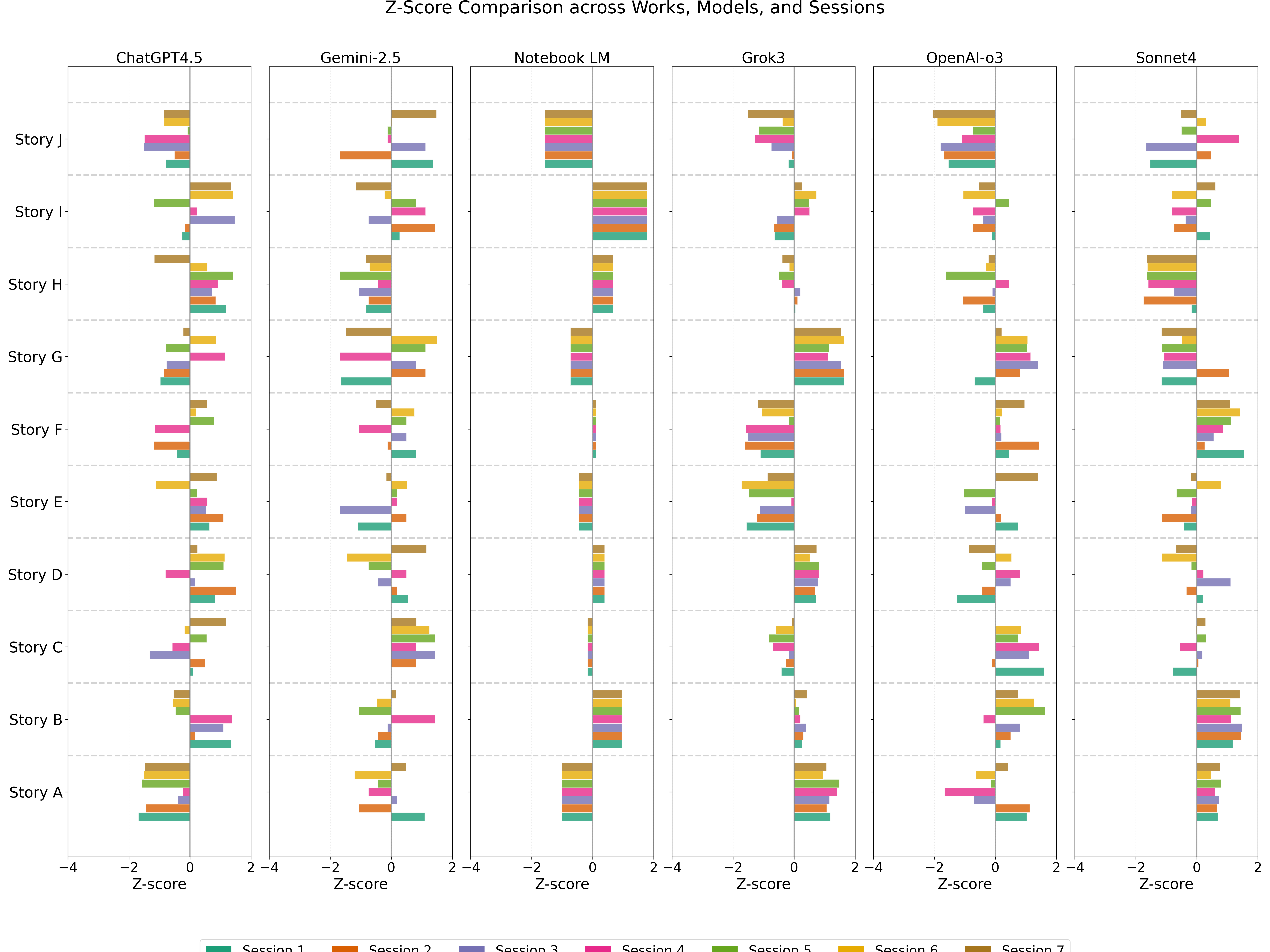}
    \caption{Evaluation consistency across models and sessions. See Appendix~\ref{appendix_F5.1} for discussion of session-timing trade-offs.}
    \label{fig_sessionized_Z}
\end{figure}

A similar consistency hierarchy was confirmed using Spearman's rank correlation coefficients ($\rho$), which measure ordinal agreement across evaluation sessions. \textbf{Notebook LM} again demonstrated perfect agreement ($\rho$ = 1.00), followed by \textbf{Grok3} ($\rho$ = 0.83). In contrast, \textbf{Gemini-2.5} showed almost no correlation across sessions ($\rho$ = 0.10). These patterns are visually evident in the Z-score distributions by model and session shown in Figure~\ref{fig_sessionized_Z}.

To statistically validate these patterns, we conducted multiple pairwise comparisons with FDR correction. Only \textbf{Notebook LM} and \textbf{Grok3} demonstrated consistent detection of significant differences across all pairwise tests (100\% and 77.8\% detection rates, respectively). Although \textbf{Sonnet4} showed relatively high $\alpha$ (0.89), its post-FDR detection rate dropped to 53.3\%, highlighting a potential gap between internal consistency and discriminative power. This analysis confirms that models exhibiting higher consistency metrics are more likely to detect statistically significant distinctions in literary evaluation.

\subsection{Inter-Story Differences in Evaluation Variance}
Large differences in evaluation variance were observed between stories (story-specific detailed statistics in Appendix~\ref{appendix_A1}):
\begin{figure}[H]
    \includegraphics[width=\linewidth]{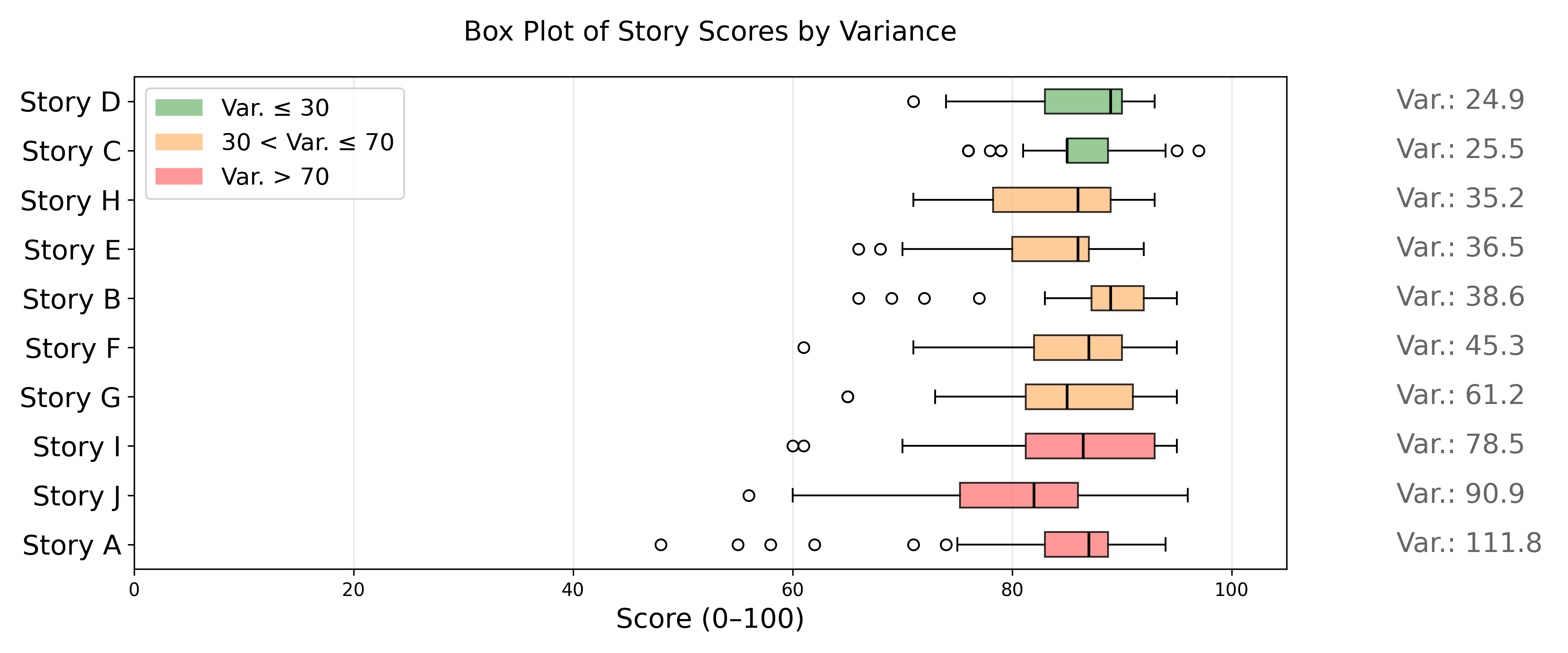}
    \caption{Actual score distribution by story}
    \label{fig_boxplot_act}
\end{figure}

Based on the variance-based color coding in Figure \ref{fig_boxplot_act}, the three high-variance stories (Var. > 70) are:
\begin{itemize}
    \item Story A (111.8), Story J (90.9), Story I (78.5)
\end{itemize}

The three low-variance stories (Var. $\leq$ 30) are:
\begin{itemize}
    \item Story D (24.9), Story C (25.5), and none other qualified (only two fall in this range).
\end{itemize}

\begin{figure}[H]
    \includegraphics[width=\linewidth]{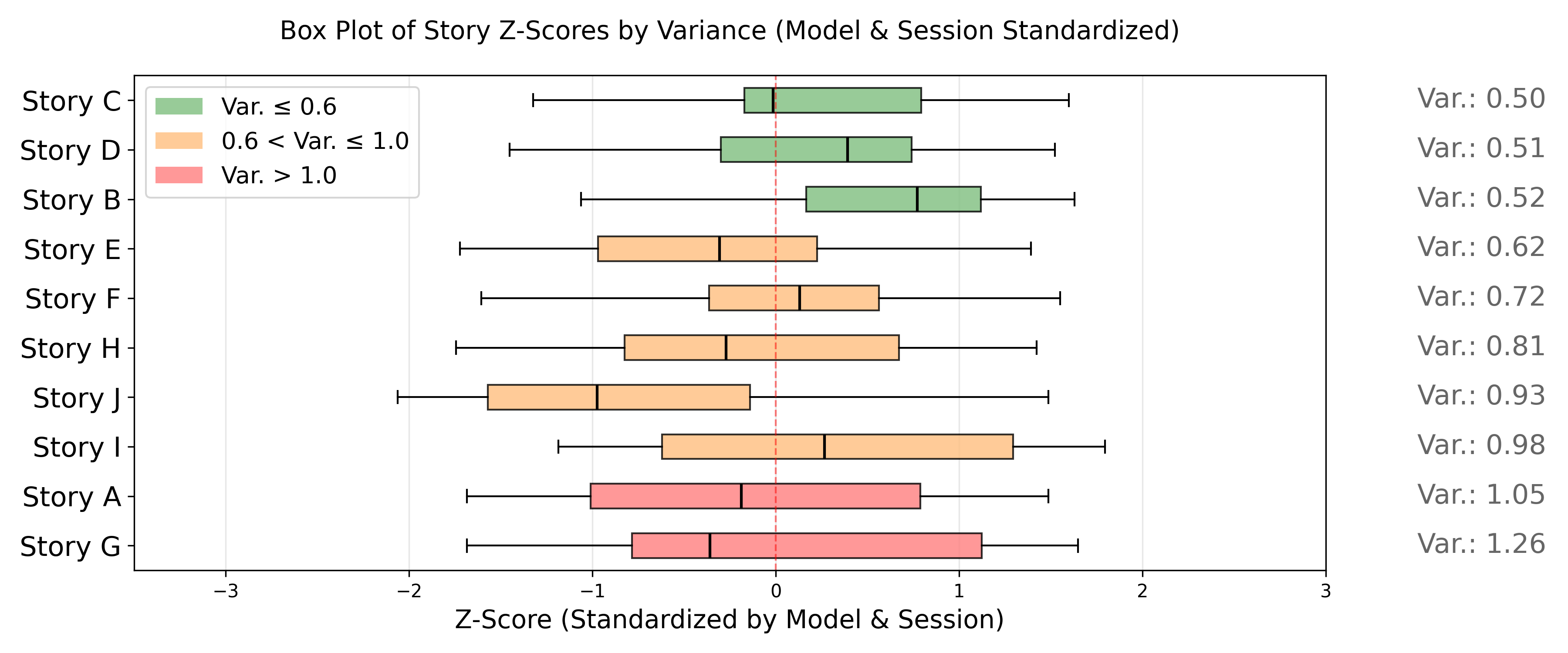}
    \caption{Z-score distribution by story}
    \label{fig_boxplot_Z}
\end{figure}

As shown in Figure \ref{fig_boxplot_Z}, a substantial 4.5-fold difference was observed between the highest and lowest Z-score variances, indicating significant variability in relative evaluation rankings across models and sessions. Story G (variance: 1.26) and Story A (variance: 1.05) exhibited the largest positional fluctuations, both belonging to the high-variance category (Var. > 1.0), represented in red. In contrast, Story C (0.50), Story D (0.51), and Story B (0.52) fell into the low-variance group (Var. ≤ 0.6), marked in green, suggesting consistent evaluations across sessions. These findings underscore that while some stories elicited stable reader impressions, others provoked divergent interpretations depending on model and trial.

\subsection{Relationship Between Evaluation Variance and Story Features}
Here we preliminarily explore evaluation variance differences between stories and potential factors. The large differences in evaluation variance between stories (up to 4.5-fold) make it important to explore factors behind this phenomenon. Future quantitative analysis linking textual features remains a research challenge. At present, the possibility of linguistic statistical differences between high and low variance stories is suggested.

\subsection{Evaluation Language Analysis}
\label{sec_4.4}
TF-IDF analysis of each model's evaluation comments identified language patterns reflecting different "evaluation profiles" (detailed vocabulary analysis in Appendix~\ref{appendix_E}):
\begin{itemize}
    \item \textbf{ChatGPT-4.5} featured words like "emotional," "resonance," "cohesion" at the top, revealing constructive evaluation tendencies emphasizing harmony between emotion and logic.
    \item \textbf{Gemini-2.5} prominently featured vocabulary like "human," "concept," "unsettling," showing evaluation tendencies exploring human existence boundaries and creativity.
    \item \textbf{Notebook LM} evaluation language featured "human," "consciousness," "psychological" at the top, showing strong orientation toward philosophical and psychological depth exploration.
    \item \textbf{OpenAI-o3} most characteristically featured "emotional," "exposition," "prose," clearly showing evaluation tendencies emphasizing stylistic beauty that evokes emotion.
    \item \textbf{Grok3} characteristically featured vocabulary like "emotional," "literary," "concept," "aesthetic," reflecting harmonious evaluation approaches emphasizing literary resonance.
    \item \textbf{Sonnet4} had "emotional," "human," "philosophical," "feels" at the top, confirming tendencies to place philosophy and emotion resonance as evaluation axes.
\end{itemize}

Evaluation vocabulary differences confirmed by TF-IDF analysis (Appendix Table~\ref{table_TF-IDF_model}) indicate that each model uses different evaluation frameworks and language repertoires.

TF-IDF analysis was also conducted on the five clusters identified in Section 4.6 to explore linguistic features of evaluation profiles transcending models. However, considering large sample size differences between clusters (30 to 130 cases) and bias from Notebook LM's complete overlap, this analysis was positioned as supplementary. Analysis results showed "originality" and "humanity" as characteristic words in Cluster 1, and "philosophical" and "consciousness" in Cluster 4, confirming linguistic differentiation of evaluation profiles (details in Appendix Table~\ref{table_TF-IDF_cluster}).

\subsection{Principal Component Analysis Results}
Principal component analysis extracted major dimensions explaining AI models' evaluation profiles (detailed factor loadings in Appendix~\ref{appendix_C}). The first four principal components cumulatively explained 76.32\% of the total variance: PC1 (28.15\%), PC2 (19.42\%), PC3 (16.39\%), and PC4 (12.37\%). Notably, the first two components alone accounted for 47.57\% of the variance, indicating a strong two-dimensional structure in the evaluation differences between AI models.

\begin{figure}[H]
    \includegraphics[width=\linewidth]{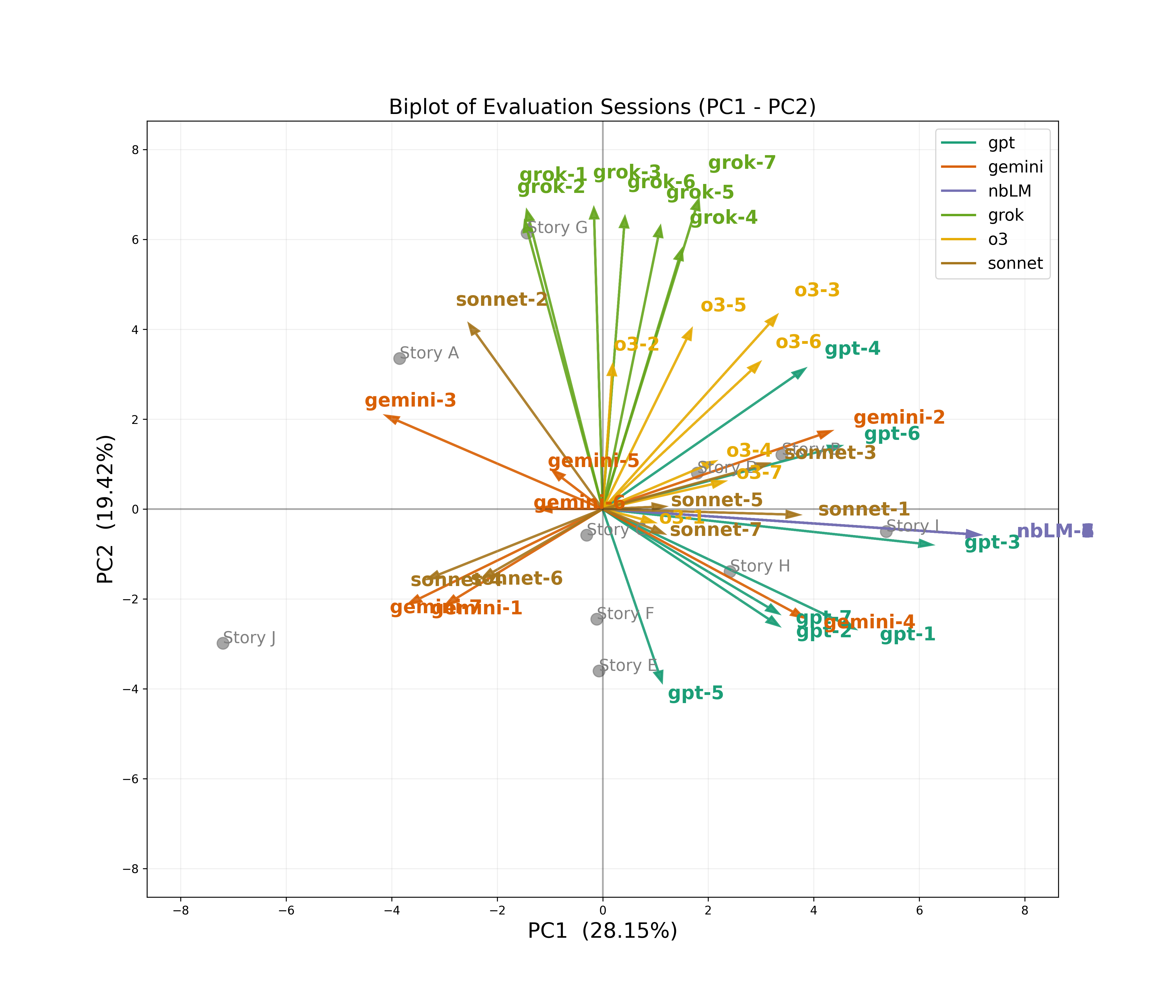}
    \caption{Principal component analysis results}
    \label{fig_PCA_biplot}
\end{figure}

From AI model arrangement patterns in the PC1-PC2 plane (Figure \ref{fig_PCA_biplot}), the following features are observed:

\begin{enumerate}
    \item \textbf{Separation structure by axes}: Notebook LM (all 7 evaluations overlapping) separated in positive direction of PC1 axis, with some Gemini-2.5 separated in negative direction. On PC2 axis, many Grok3 evaluations and Sonnet4's Sonnet-2 occupied extreme positions in positive direction, while ChatGPT-4.5 and some Gemini-2.5 were positioned in negative direction.
    \item \textbf{Model-specific variance characteristics}: Grok3 and Sonnet4 showed the widest variance in principal component space, with Sonnet-2 particularly positioned far from all other evaluations. Meanwhile, Notebook LM showed zero variance due to perfect overlap, while OpenAI-o3 showed moderate variance patterns.
\end{enumerate}

\subsection{AI Model Evaluation Clustering}
Clustering analysis using principal component scores from PC1 to PC4 confirmed that AI model evaluation profiles are classified into clear groups (clustering method details in Appendix~\ref{appendix_D}).

Multiple indicators were comprehensively considered for determining cluster number. Silhouette score analysis observed maximum values for 4 clusters (K-means: 0.420, hierarchical: 0.402), but considering natural division points in hierarchical clustering dendrograms (Appendix Figure~\ref{fig_dendrogram}) and interpretability, we ultimately adopted 5 clusters. Silhouette scores for 5 clusters also maintained high values (K-means: 0.418, hierarchical: 0.417), ensuring statistical validity.

\begin{figure}[H]
    \includegraphics[width=\linewidth]{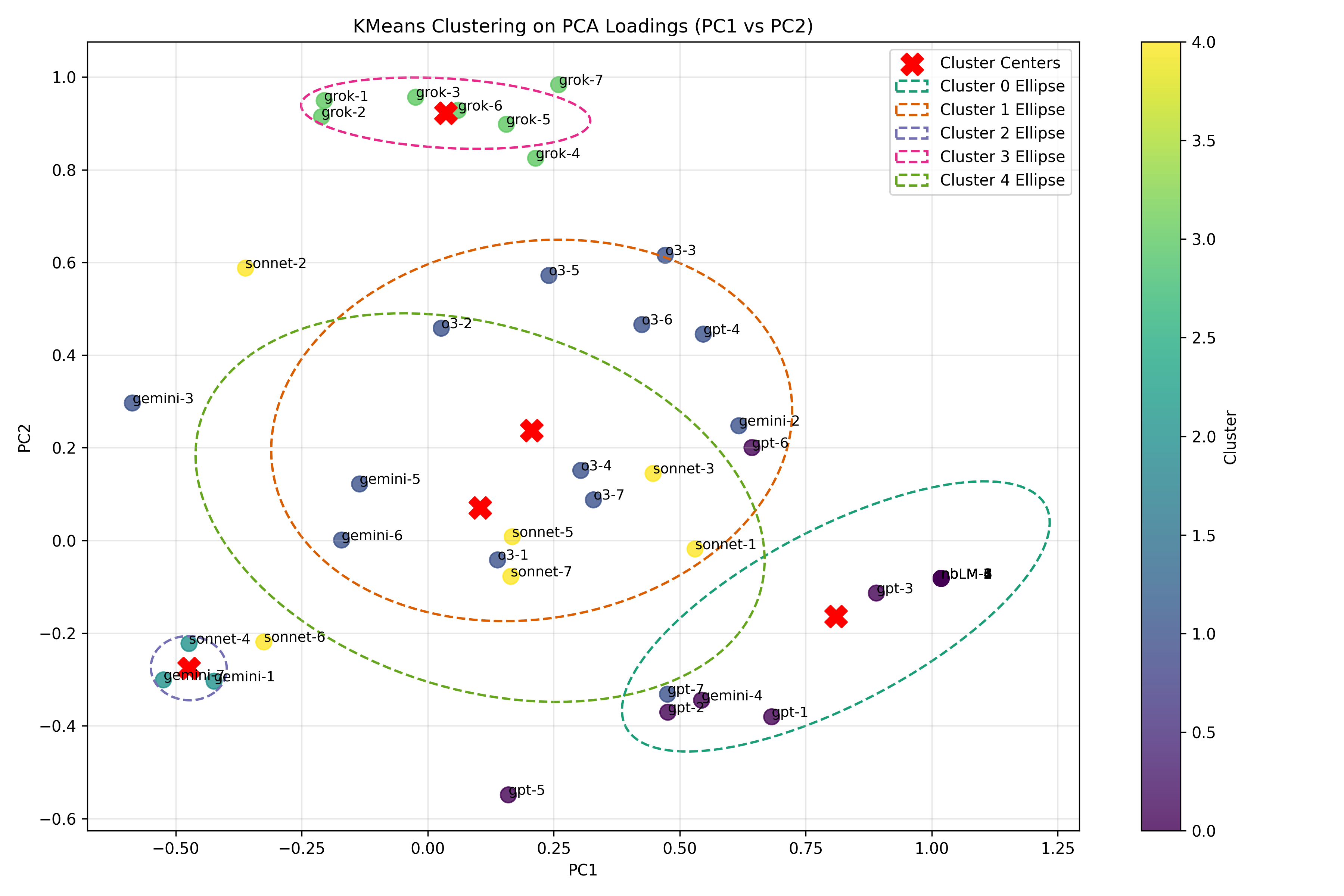}
    \caption{Clustering results}
    \label{fig_kmeans}
\end{figure}

K-means method classified 42 AI model evaluations into five clear clusters (Figure \ref{fig_kmeans}). Overall average silhouette score was 0.418, confirming statistically significant cluster structure.
Results from hierarchical clustering method also showed high agreement with K-means method, supporting the robustness of discovered evaluation profiles (detailed individual model analysis in Appendix~\ref{appendix_D2}).

\section{Discussion}

\subsection{Diversity of AI Model Evaluation Profiles}
The four major design trade-offs (Tables presented in Appendix~\ref{appendix_F}) frame our interpretation of variance. Statistical analysis in this study demonstrated that LLMs exhibit statistically distinguishable characteristics in literary evaluation. The hierarchical structure of evaluation consistency ($\alpha$ ranging from 1.00 to 0.35) indicates fundamental differences in evaluation processes between models.

Particularly, Notebook LM's perfect consistency (α=1.00) suggests deterministic evaluation processes, while Gemini-2.5's low consistency (α=0.35) indicates large variation between evaluation sessions. The approximately three-fold difference observed from Grok3's high consistency (α=0.96) to Gemini-2.5's low consistency suggests that evaluation approach stability varies greatly between models even for identical literary works. This phenomenon suggests the possibility that different evaluation frameworks may be activated depending on evaluation sessions even within the same model.

\subsection{Relationship Between Evaluation Variance and Story Features}
From objective data analysis, stories showed characteristic arrangement patterns in principal component space. Future quantitative analysis linking textual features is expected to identify factors promoting evaluation diversity (research limitations and future prospects in Appendix~\ref{appendix_F3}).

\subsection{Structure in Principal Component Space}
The four-dimensional structure extracted by principal component analysis (cumulative contribution rate 76.32\%) indicates that AI model evaluation behavior can be characterized by relatively few basic dimensions. Particularly, the fact that PC1 and PC2 alone explain 47.57\% of variance suggests the existence of fundamental two-dimensional structure in evaluation profiles.

Given the SF-only corpus (Appendix~\ref{appendix_F}), the PCA axes should be interpreted as genre-internal rather than cross-genre dimensions. See Appendix~\ref{appendix_F5.2} for a detailed discussion of corpus-choice trade-offs. The contrasting pattern of Notebook LM's complete overlap positioning (all 7 at identical positions) and Grok3/Sonnet4's wide dispersion visually demonstrates fundamental differences in evaluation strategies between models. Particularly, the fact that Sonnet4's Sonnet-2 occupies an extreme position far from all other evaluations suggests the possibility that dramatically different evaluation dimensions may be activated depending on evaluation sessions even within a single model.

\subsection{Evaluation Profiles Through Clustering}
The statistical validity of five-cluster composition (silhouette score 0.417-0.418) and high agreement between K-means and hierarchical methods indicate the existence of natural typologies in AI model evaluation profiles. Cluster composition is potentially sensitive to prompt conditioning (Appendix~\ref{appendix_F}); future work will test stability under alternative critical prompts.

Particularly, the pattern where all Grok3 evaluations form a single cluster (k-means cluster 3) while clusters with mixed model families exist suggests that evaluation characteristics may be determined by factors transcending model architecture. The fact that Sonnet4 and OpenAI-o3 distribute across multiple clusters means these models exhibit different evaluation profiles depending on evaluation sessions.

Cluster-specific TF-IDF analysis (supplementary experiment) confirmed linguistic features of evaluation profiles transcending model architecture. For example, "originality" and "humanity" appeared as characteristic words in Cluster 1, while "philosophical" and "consciousness" in Cluster 4, with evaluation convergence phenomena observed at vocabulary level. This suggests the possibility that models from different development companies activate common evaluation frameworks in specific evaluation contexts.

\subsection{Evaluation Language Differentiation}
Evaluation vocabulary differences confirmed by TF-IDF analysis indicate that each model uses different evaluation frameworks and language repertoires (See Appendix Table~\ref{table_TF-IDF_model}). This suggests that evaluation score pattern differences may not be superficial but reflect deeper differences in evaluation cognitive processes.

\subsection{Theoretical Implications}
Our study's findings indicate the need to understand AI model evaluation behavior not as processes seeking single "correct answers" but as processes providing multiple valid perspectives. This suggests the possibility of realizing "readers' active roles" from reader-response theory (Iser 1978\cite{Iser1978}) in AI contexts.

Additionally, this provides quantitative evidence supporting Bode (2023)\cite{Bode2023}'s perspective of accepting model subjectivity. Our results thus align with the \textit{AI value-system archaeology} research agenda sketched in Appendix~\ref{appendix_F5}.

\section{Conclusion}

\subsection{Major Findings}
\begin{enumerate}
    \item \textbf{Diversity of evaluation processes}: Statistical analysis confirmed that AI model evaluation differences possess structured diversity. The 76.32\% variance explanation rate from principal component analysis and significant cluster structure in 5 clusters indicate this diversity is not random variation.
    \item \textbf{Hierarchy of evaluation consistency}: Clear hierarchical structure was confirmed from perfect consistency (Notebook LM, $\alpha$=1.00) to low consistency (Gemini-2.5, $\alpha$=0.35). Particularly, Grok3's high consistency ($\alpha$=0.96) and Sonnet4's stable consistency ($\alpha$=0.89) are positioned as new findings in the six-model framework.
    \item \textbf{Evaluation variance differences by story characteristics}: Up to 4.5-fold differences in evaluation variance were observed between stories, confirming that specific stories promote evaluation divergence between models. This phenomenon suggests the existence of story intrinsic feature influences on evaluator judgment processes.
    \item \textbf{Supplementary exploratory results (cluster-specific TF-IDF)}: Vocabulary patterns converging within clusters across models were confirmed. Examples: Cluster 1: originality/humanity, Cluster 4: philosophical/consciousness. This suggests cross-cutting evaluation profile linguistic aspects not revealed in single-model analysis (details in Section \ref{sec_4.4}). This analysis remains exploratory verification due to imbalanced text volumes between clusters.
\end{enumerate}

\subsection{Academic Contributions}
This study established an integrated analytical framework for AI literary evaluation research, demonstrating that AI model evaluation processes should be understood as probabilistic distributions. This indicates a theoretical shift from deterministic to probabilistic evaluation perspectives.

\subsection{Methodological Significance}
Through mutual verification approaches using multiple statistical methods, we presented methodology for objectively measuring and analyzing AI subjective evaluation behavior. Particularly, the combination of evaluation consistency quantification and evaluation language quantitative analysis can become important methodology for future research.

\subsection{Limitations and Prospects}
Study limitations include sample size constraints (6 models × 7 evaluations × 10 works), causal inference limitations from observational design, and restriction to specific time points and genres. Future research requires verification with larger-scale samples, experimental designs, and diverse genres (details in Appendix~\ref{appendix_F}).

\subsection{Final Conclusion}
This study established the fundamental recognition that AI models possess diverse evaluation characteristics rather than single evaluation criteria, supported by quantitative evidence. This finding presents a new paradigm positioning AI literary evaluation as "providing diverse perspectives" rather than "objective judgment."

These findings, demonstrated through statistically robust analysis, provide theoretical foundations for AI literary evaluation research while opening new possibilities for AI-assisted literary understanding.

\clearpage
\appendix

\section{APPENDIX - Basic Statistics of Experimental Data}
\subsection{Basic Story Information and Evaluation Statistics}
\label{appendix_A1}

\newcolumntype{Y}{>{\centering\arraybackslash}X}
\begin{table}[ht]
    \centering
    \begin{tabularx}{\linewidth}{l Y Y Y Y Y Y}
    \toprule
    \thead{\textbf{Story}} & 
    \thead{\textbf{Word}\\\textbf{Count}} & 
    \thead{\textbf{Mean}\\\textbf{Score}} & 
    \thead{\textbf{Std Dev}} & 
    \thead{\textbf{Raw Score}\\\textbf{Variance}} & 
    \thead{\textbf{Z-Score}\\\textbf{Variance}} & 
    \thead{\textbf{Range}} \\
    \midrule
    Story A & 2,291 & 83.3 & 10.57 & 111.8 & 1.05 & 48-94 \\
    Story B & 3,832 & 87.8 & 6.21 & 38.6 & 0.52 & 66-95 \\
    Story C & 1,962 & 86.1 & 5.05 & 25.5 & 0.50 & 76-97 \\
    Story D & 1,671 & 86.6 & 4.99 & 24.9 & 0.51 & 71-93 \\
    Story E & 1,998 & 83.7 & 6.04 & 36.5 & 0.62 & 66-92 \\
    Story F & 3,225 & 85.4 & 6.73 & 45.3 & 0.72 & 61-95 \\
    Story G & 3,068 & 85.4 & 7.82 & 61.2 & 1.26 & 65-95 \\
    Story H & 1,279 & 83.9 & 5.94 & 35.2 & 0.81 & 71-93 \\
    Story I & 2,145 & 85.2 & 8.86 & 78.5 & 0.98 & 60-95 \\
    Story J & 1,765 & 80.0 & 9.53 & 90.9 & 0.93 & 56-96 \\
    \bottomrule
    \end{tabularx}
    \caption{Basic story information and evaluation statistics}
\end{table}

\subsection{Model-specific Evaluation Consistency Indicators}
\label{appendix_A2}
\newcolumntype{Y}{>{\centering\arraybackslash}X}
\begin{table}[ht]
    \centering
    \begin{tabularx}{\linewidth}{l Y Y Y Y}
    \toprule
    \thead{\textbf{Model}} &
    \thead{\textbf{Cronbach's $\alpha$}} &
    \thead{\textbf{Mean}\\\textbf{Correlation ($\rho$)}} &
    \thead{\textbf{Top 3}\\\textbf{Agreement}} &
    \thead{\textbf{Bottom 3}\\\textbf{Agreement}} \\
    \midrule
    Notebook LM   & 1.00 & 1.00 & 100.0\% & 100.0\% \\
    Grok3         & 0.96 & 0.83 & 95.2\%  & 100.0\% \\
    Sonnet4       & 0.89 & 0.55 & 81.0\%  & 95.2\% \\
    OpenAI-o3     & 0.81 & 0.40 & 95.2\%  & 95.2\% \\
    ChatGPT-4.5   & 0.69 & 0.28 & 90.5\%  & 76.2\% \\
    Gemini-2.5    & 0.35 & 0.10 & 85.7\%  & 85.7\% \\
    \bottomrule
    \end{tabularx}
    \caption{Model-specific evaluation consistency statistics}
\end{table}

\begin{figure}[H]
    \centering
    \includegraphics[width=\linewidth]{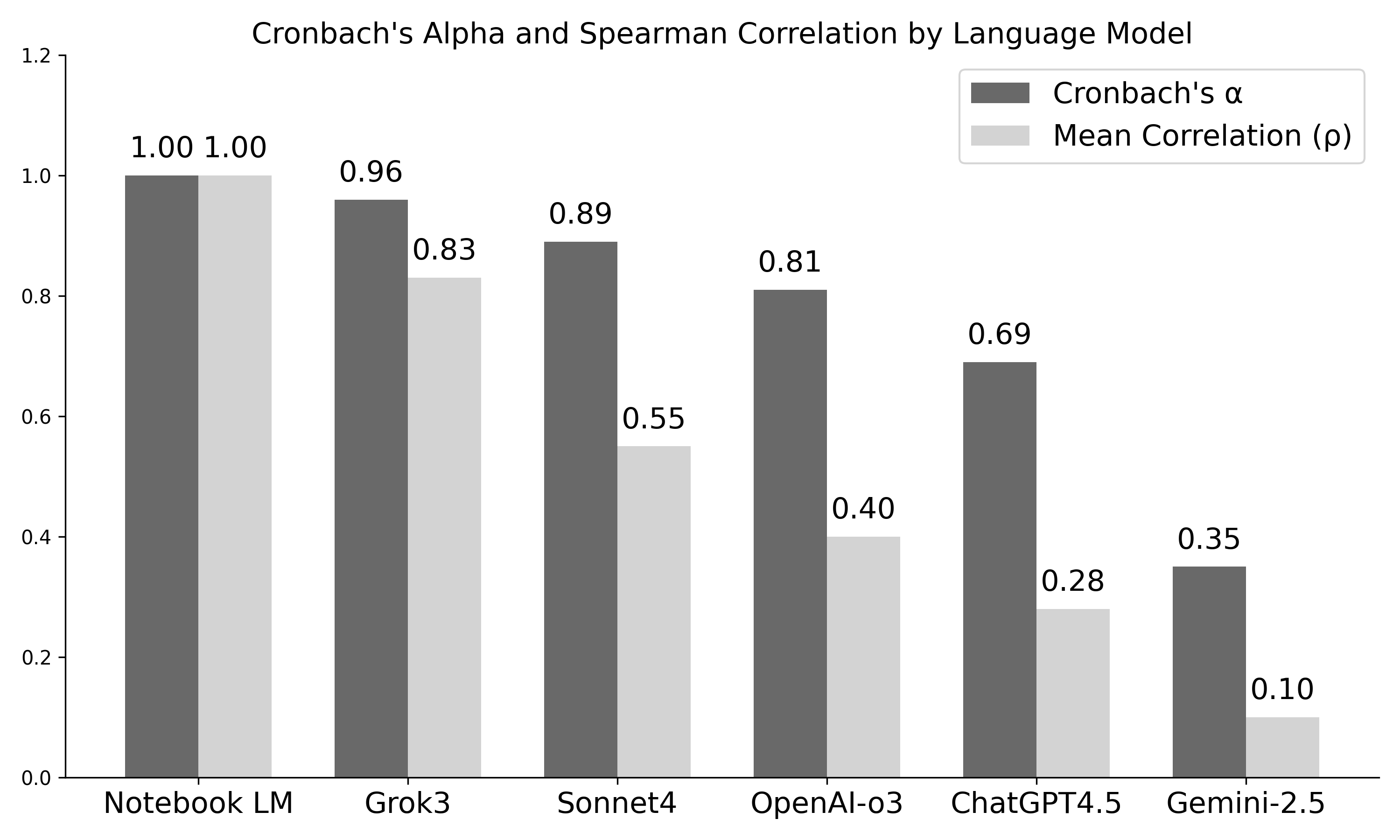}
    \caption{Inter-session internal consistency metrics}
\end{figure}

\section{Appendix - Evaluation Protocol Details}
\label{appendix_B}

\subsection{Standardized Evaluation Prompt}
\begin{tcolorbox}[colback=gray!5, colframe=black, title=Standardized Evaluation Prompt Used in Experiment, fonttitle=\bfseries]

You are a literary critic. Please evaluate the following ten English-language short stories. For each story, rate it on a 100-point scale using the following four criteria:

    \begin{enumerate}
        \item Narrative structure and cohesion
        \item Originality of the concept or theme  
        \item Emotional and aesthetic resonance
        \item Overall literary quality
    \end{enumerate}
    
    \textbf{Critical Evaluation Requirements:}
    \begin{itemize}
        \item Each story must receive a different score. No tied scores are permitted.
        \item Even if stories are closely matched in quality, you must determine relative superiority and assign distinct scores.
        \item The difference between your highest and lowest scores must be at least 10 points.
    \end{itemize}
    
    \textbf{Please return your evaluation in a table with the following columns:}
    \begin{itemize}
        \item Story ID (Story A, Story B, Story C, etc.)
        \item Score (0-100, all scores must be unique)
        \item Brief Comment (2-3 sentences) explaining your evaluation
    \end{itemize}

    \textbf{Note:} Stories are presented without titles to ensure unbiased evaluation based solely on content.
    
    \textbf{Important:}
    \begin{itemize}
        \item Judge each story independently based on the text content only.
        \item Do not summarize the story. Focus on your critical evaluation.
        \item Apply the same criteria consistently across all ten stories.
        \item Ensure every story receives a distinct numerical score.
    \end{itemize}
\end{tcolorbox}

\subsection{AI Model Specifications and Session Separation Methods}
\begin{itemize}
    \item Temperature and probability parameters were uncontrollable due to being private on the browser-provided UI.
    \item This study aimed to detect "evaluation profiles across models" rather than "behavioral differences due to parameters," thus adopted default settings as unified black-box conditions. Only Gemini-2.5 explicitly displayed temperature=1.0 in API response headers, which is noted as a reference value.
\end{itemize}

\renewcommand{\arraystretch}{2} 
\newcolumntype{C}[1]{>{\centering\arraybackslash}m{#1}}

\begin{table}[H]
\footnotesize
    \centering
    \begin{tabular}{|C{2cm}|C{2cm}|C{2.5cm}|C{2.5cm}|C{1.5cm}|}
    \hline
    \rule{0pt}{1.2em}\textbf{Model} & 
    \textbf{Usage Form} & 
    \textbf{Parameters} & 
    \textbf{Separation Method} &
    \textbf{Evaluation Schedule}\\
    \hline
    ChatGPT-4.5 & 
    Paid Browser & 
    Not specified / Unmodifiable & 
    Temporary Mode &
    May 23, 2025\\
    \hline
    Gemini-2.5 & 
    Vertex AI Studio & 
    Default values maintained: Temperature = 1.0, seed = 0 & 
    Standard API Separation &
    May 23, 2025\\
    \hline
    Notebook LM & 
    Free Browser & 
    Not specified / Unmodifiable & 
    New Notebook &
    May 23, 2025\\
    \hline
    OpenAI-o3 & 
    Paid Browser & 
    Not specified / Unmodifiable & 
    Temporary Mode &
    May 23, 2025\\
    \hline
    Grok3 & 
    Free Browser & 
    Not specified / Unmodifiable & 
    Ghost Mode &
    May 23, 2025\\
    \hline
    Sonnet4 & 
    Paid Browser & 
    Not specified / Unmodifiable & 
    New Chat &
    May 23, 2025\\
    \hline
    \end{tabular}
    \caption{Session separation protocol by model}
\end{table}

\section{Appendix - Detailed Principal Component Analysis Results}
\label{appendix_C}

\subsection{Principal Component Contribution Rates}
\newcolumntype{Y}{>{\centering\arraybackslash}X}
\begin{table}[H]
    \centering
    \begin{tabularx}{\linewidth}{l Y Y Y}
    \toprule
    \textbf{Component} & 
    \textbf{Eigenvalue} & 
    \thead{\textbf{Individual}\\\textbf{Contribution}} & 
    \thead{\textbf{Cumulative}\\\textbf{Contribution}} \\
    \midrule
    PC1 & 13.1383 & 28.15\% & 28.15\% \\
    PC2 & 9.0605 & 19.42\% & 47.57\% \\
    PC3 & 7.6481 & 16.39\% & 63.96\% \\
    PC4 & 5.7713 & 12.37\% & 76.32\% \\
    \bottomrule
    \end{tabularx}
    \caption{Eigenvalues and contribution rates of principal components}
\end{table}

\subsection{Factor Loadings of Evaluation Vectors in Principal Component Space}
\begin{longtable}{p{3.2cm} p{1.7cm} p{1.7cm} p{1.7cm} p{1.7cm}}
\toprule
\textbf{Evaluation ID} & 
\textbf{PC1} & 
\textbf{PC2} & 
\textbf{PC3} & 
\textbf{PC4} \\
\midrule
\endfirsthead
\multicolumn{5}{c}{\tablename\ \thetable{} -- continued from previous page} \\
\toprule
\textbf{Evaluation ID} & 
\textbf{PC1} & 
\textbf{PC2} & 
\textbf{PC3} & 
\textbf{PC4} \\
\midrule
\endhead
\midrule
\multicolumn{5}{r}{Continued on next page} \\
\endfoot
\bottomrule
\endlastfoot
    ChatGPT-4.5-1 & 0.682 & -0.380 & -0.229 & -0.044 \\
    ChatGPT-4.5-2 & 0.476 & -0.371 & -0.514 & 0.009 \\
    ChatGPT-4.5-3 & 0.890 & -0.113 & 0.075 & -0.341 \\
    ChatGPT-4.5-4 & 0.547 & 0.445 & -0.164 & 0.110 \\
    ChatGPT-4.5-5 & 0.160 & -0.549 & -0.411 & 0.196 \\
    ChatGPT-4.5-6 & 0.643 & 0.200 & -0.447 & 0.128 \\
    ChatGPT-4.5-7 & 0.476 & -0.332 & 0.172 & 0.614 \\
    \midrule
    Gemini-2.5-1 & -0.424 & -0.304 & 0.374 & -0.555 \\
    Gemini-2.5-2 & 0.617 & 0.248 & -0.040 & 0.645 \\
    Gemini-2.5-3 & -0.586 & 0.297 & 0.255 & 0.291 \\
    Gemini-2.5-4 & 0.543 & -0.345 & 0.197 & -0.267 \\
    Gemini-2.5-5 & -0.135 & 0.122 & 0.233 & 0.786 \\
    Gemini-2.5-6 & -0.171 & 0.001 & 0.085 & 0.987 \\
    Gemini-2.5-7 & -0.524 & -0.301 & 0.140 & -0.371 \\
    \midrule
    Notebook LM-1 & 1.019 & -0.082 & 0.047 & -0.146 \\
    Notebook LM-2 & 1.019 & -0.082 & 0.047 & -0.146 \\
    Notebook LM-3 & 1.019 & -0.082 & 0.047 & -0.146 \\
    Notebook LM-4 & 1.019 & -0.082 & 0.047 & -0.146 \\
    Notebook LM-5 & 1.019 & -0.082 & 0.047 & -0.146 \\
    Notebook LM-6 & 1.019 & -0.082 & 0.047 & -0.146 \\
    Notebook LM-7 & 1.019 & -0.082 & 0.047 & -0.146 \\
    \midrule
    Grok3-1 & -0.205 & 0.949 & -0.218 & -0.281 \\
    Grok3-2 & -0.211 & 0.914 & -0.325 & -0.256 \\
    Grok3-3 & -0.024 & 0.957 & -0.260 & -0.226 \\
    Grok3-4 & 0.214 & 0.825 & -0.105 & -0.287 \\
    Grok3-5 & 0.156 & 0.898 & 0.147 & -0.340 \\
    Grok3-6 & 0.060 & 0.928 & -0.194 & -0.294 \\
    Grok3-7 & 0.260 & 0.984 & -0.041 & -0.081 \\
    \midrule
    OpenAI-o3-1 & 0.138 & -0.042 & 0.600 & 0.394 \\
    OpenAI-o3-2 & 0.027 & 0.458 & 0.706 & 0.312 \\
    OpenAI-o3-3 & 0.471 & 0.615 & 0.079 & 0.493 \\
    OpenAI-o3-4 & 0.304 & 0.151 & -0.369 & 0.734 \\
    OpenAI-o3-5 & 0.240 & 0.572 & 0.609 & 0.278 \\
    OpenAI-o3-6 & 0.425 & 0.466 & 0.218 & 0.494 \\
    OpenAI-o3-7 & 0.329 & 0.088 & 0.525 & 0.417 \\
    \midrule
    Sonnet4-1 & 0.530 & -0.018 & 0.704 & -0.346 \\
    Sonnet4-2 & -0.362 & 0.588 & 0.606 & 0.010 \\
    Sonnet4-3 & 0.447 & 0.145 & 0.660 & -0.267 \\
    Sonnet4-4 & -0.474 & -0.222 & 0.679 & -0.397 \\
    Sonnet4-5 & 0.168 & 0.008 & 0.990 & -0.207 \\
    Sonnet4-6 & -0.325 & -0.219 & 0.893 & 0.177 \\
    Sonnet4-7 & 0.165 & -0.078 & 1.017 & -0.116 \\
\caption{Principal component factor loadings of evaluation vectors}
\end{longtable}

\subsection{Story Coordinates in Principal Component Space}
\newcolumntype{Y}{>{\centering\arraybackslash}X}
\begin{table}[ht]
    \centering
    \begin{tabularx}{\linewidth}{l Y Y Y Y}
    \toprule
    \textbf{Story} & 
    \textbf{PC1} & 
    \textbf{PC2} & 
    \textbf{PC3} & 
    \textbf{PC4} \\
    \midrule
    Story A & -3.850 & 3.350 & 2.572 & -3.047 \\
    Story B & 3.393 & 1.212 & 3.530 & -1.624 \\
    Story C & -0.303 & -0.582 & 0.732 & 3.597 \\
    Story D & 1.793 & 0.803 & -2.001 & -2.101 \\
    Story E & -0.072 & -3.603 & -0.252 & 2.017 \\
    Story F & -0.114 & -2.449 & 3.841 & 1.652 \\
    Story G & -1.434 & 6.147 & -2.181 & 3.295 \\
    Story H & 2.412 & -1.392 & -4.853 & -1.189 \\
    Story I & 5.374 & -0.500 & 0.059 & -0.966 \\
    Story J & -7.199 & -2.986 & -1.447 & -1.634 \\
    \bottomrule
    \end{tabularx}
    \caption{Story coordinates in principal component space}
\end{table}

\begin{figure}[H]
    \centering
    \includegraphics[width=\linewidth]{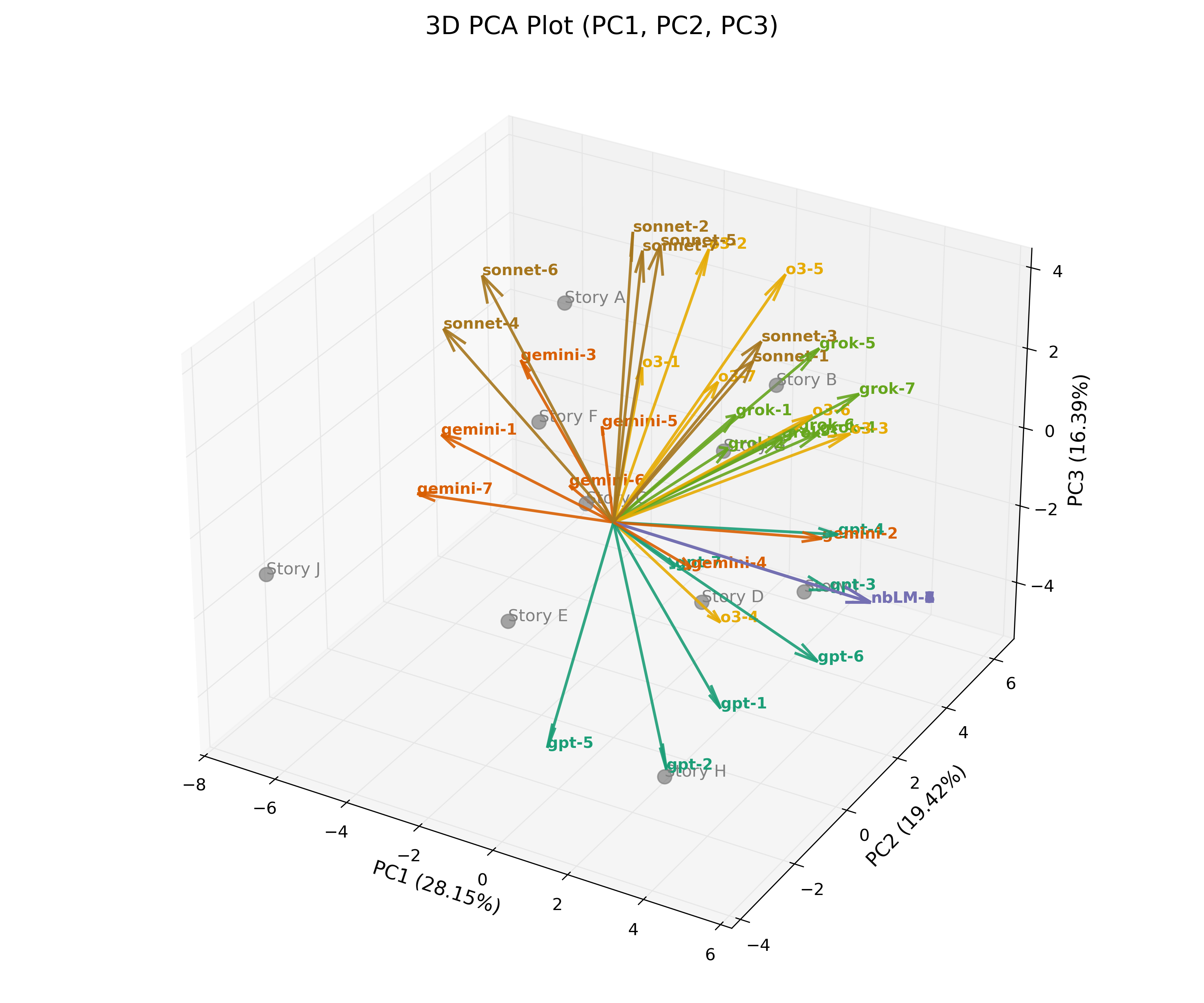}
    \caption{Evaluation distribution in 3d principal component space}
\end{figure}
\textit{Note: The 3D representation shows complex relationships not captured in the biplot of Section~\ref{sec_4.4}. Colors represent model families.}

\section{Appendix - Detailed Clustering Analysis}
\label{appendix_D}
\subsection{K-means Clustering Results}
\newcolumntype{Y}{>{\centering\arraybackslash}X}
\begin{table}[ht]
    \centering
    \begin{tabularx}{\linewidth}{Y Y Y}
    \toprule
    \textbf{Cluster Number} & 
    \textbf{Silhouette Score} & 
    \textbf{WSS} \\
    \midrule
    2 & 0.290 & 20.438 \\
    3 & 0.334 & 15.011 \\
    4 & 0.420 & 10.051 \\
    5 & 0.418 & 8.646 \\
    6 & 0.407 & 7.416 \\
    7 & 0.410 & 6.169 \\
    \bottomrule
    \end{tabularx}
    \caption{Cluster number evaluation indicators in k-means clustering}
\end{table}

\begin{figure}[H]
    \centering
    \includegraphics[width=\linewidth]{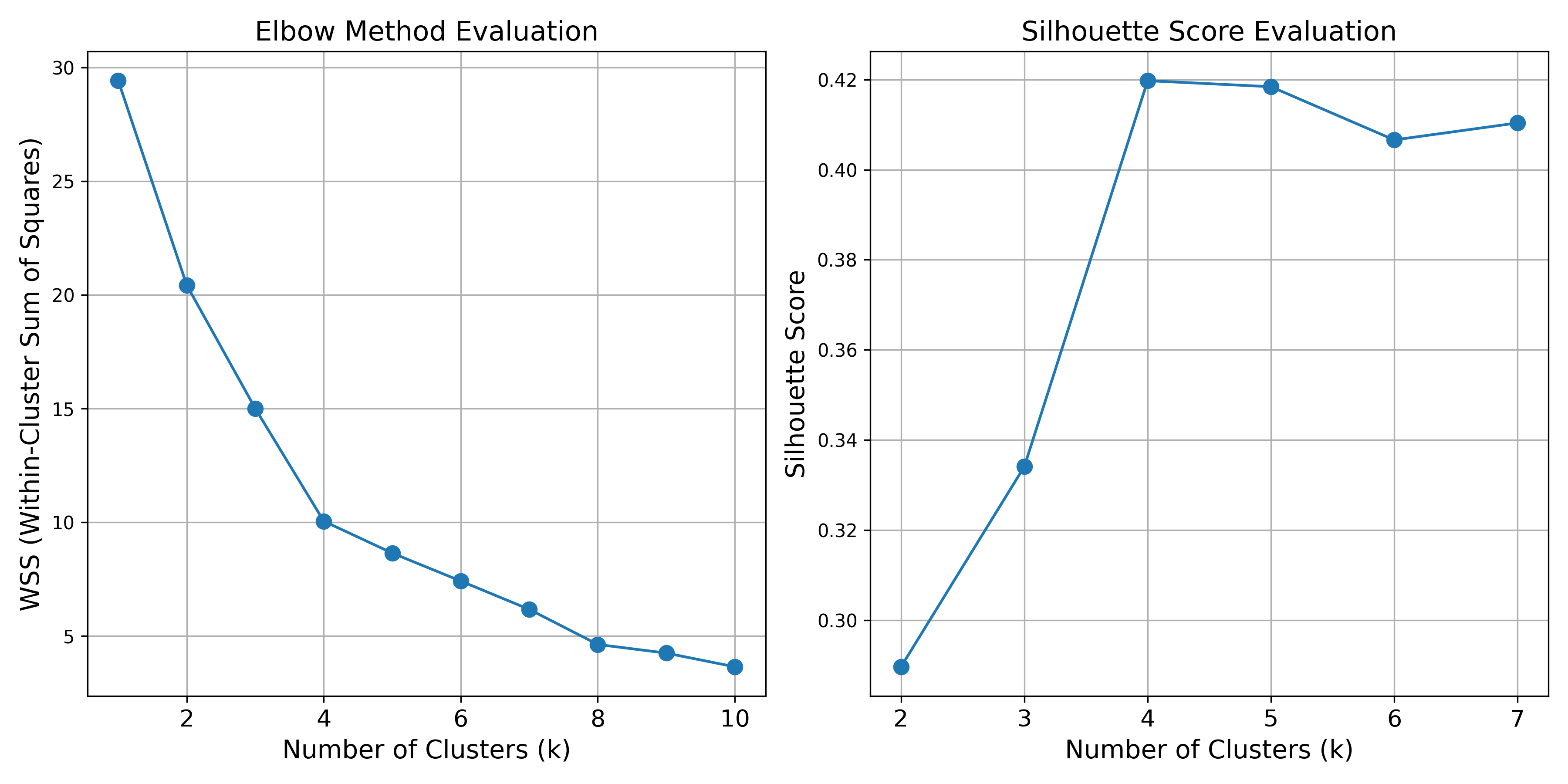}
    \caption{Cluster number evaluation plots in l-means clustering}
\end{figure}

\subsection{Hierarchical Clustering Results}
\label{appendix_D2}
\newcolumntype{Y}{>{\centering\arraybackslash}X}
\begin{table}[ht]
    \centering
    \begin{tabularx}{\linewidth}{Y Y Y}
    \toprule
    \textbf{Cluster Number} & 
    \textbf{Silhouette Score} & 
    \textbf{Within-Cluster Sum of Squares (WSS)} \\
    \midrule
    2 & 0.283 & 22.573 \\
    3 & 0.360 & 14.741 \\
    4 & 0.402 & 10.390 \\
    5 & 0.417 & 8.385 \\
    6 & 0.417 & 7.299 \\
    7 & 0.421 & 6.122 \\
    \bottomrule
    \end{tabularx}
    \caption{Cluster number evaluation indicators in hierarchical clustering}
\end{table}

\begin{figure}[H]
    \centering
    \includegraphics[width=\linewidth]{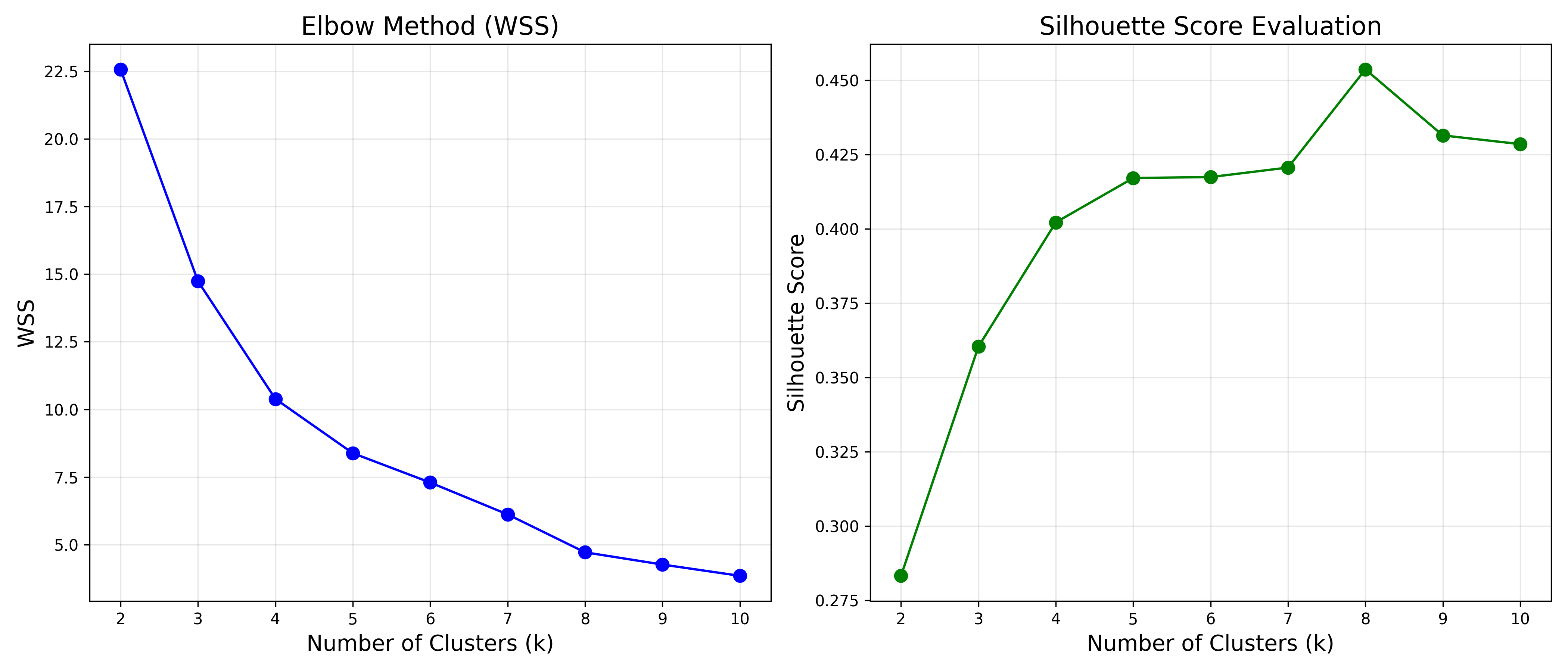}
    \caption{Cluster number evaluation plots in hierarchical clustering}
\end{figure}

\begin{figure}[H]
    \centering
    \includegraphics[width=\linewidth]{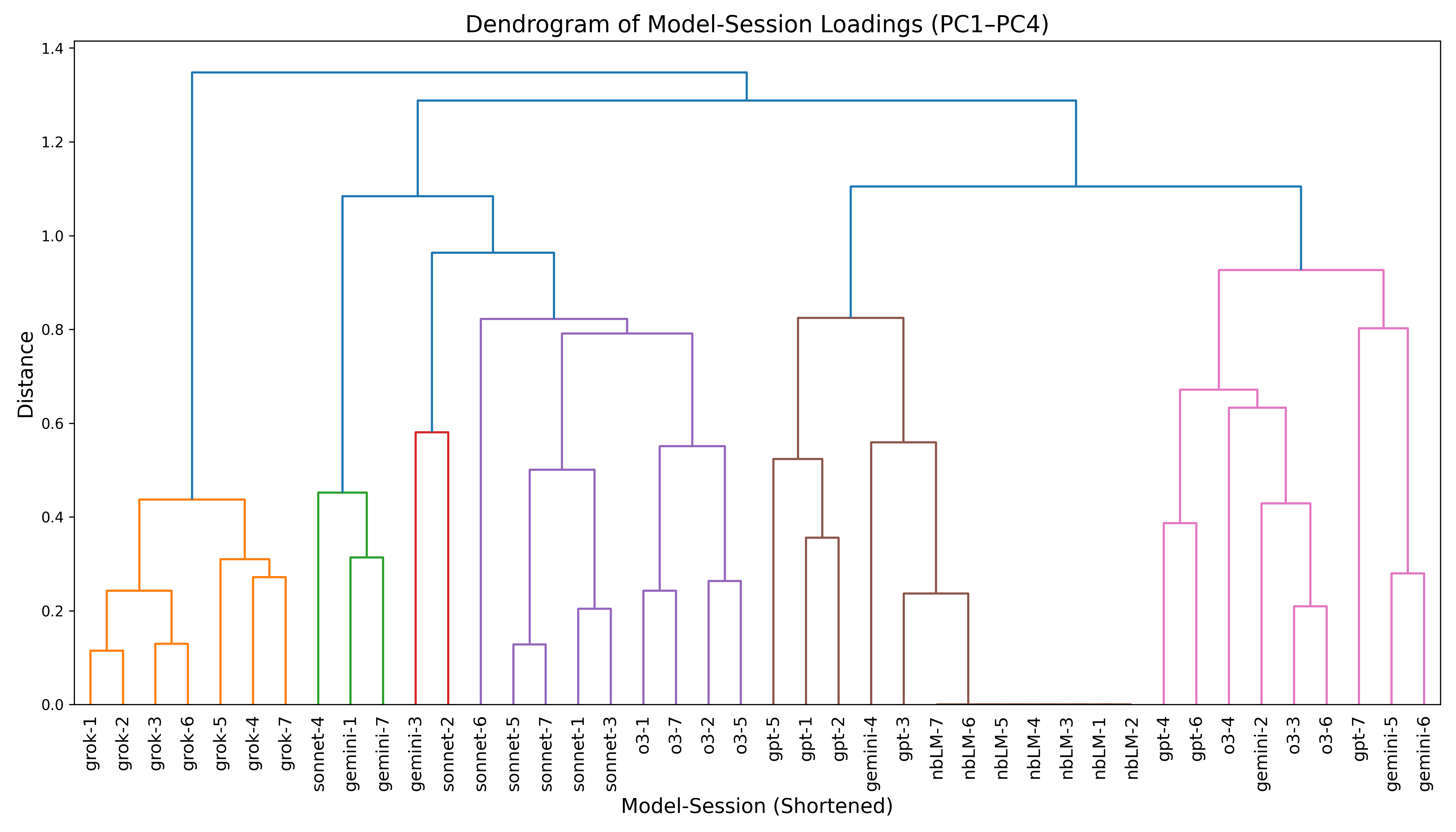}
    \caption{Dendrogram showing hierarchical clustering results}
    \label{fig_dendrogram}
\end{figure}
\textit{Note: Vertical distances represent dissimilarity in evaluation profiles, and the horizontal red line indicates division at the 5-cluster level.}

\subsection{Detailed Results of Multiple Comparison Correction}
\label{appendix_D3}
\subsubsection{Overview of FDR Correction Implementation}
To control Type I error from multiple comparisons, we applied FDR correction using the Benjamini-Hochberg method ($\alpha$ = 0.05) for each model's inter-session evaluation consistency. The analysis targets were as follows:

\begin{itemize}
    \item \textbf{Pairwise t-tests}: Inter-session comparisons within each model (45 comparisons: 10 stories $\times$ C(7,2))
    \item \textbf{Spearman rank correlation tests}: Significance testing of rank correlations within each model (21 comparisons: C(7,2))
\end{itemize}

\subsubsection{Re-evaluation of Statistical Significance by FDR Correction}
\newcolumntype{Y}{>{\centering\arraybackslash}X}
\begin{table}[ht]
    \centering
    \begin{tabularx}{\linewidth}{Y Y Y Y}
    \toprule
    \textbf{Analysis Type} & 
    \textbf{Model} & 
    \thead{\textbf{Significant}\\\textbf{Differences/Total}\\\textbf{(q $<$ 0.05)}} & 
    \thead{\textbf{Detection}\\\textbf{Rate (\%)}\\\textbf{[Post-FDR}\\\textbf{Correction]}} \\
    \midrule
    \rule{0pt}{10pt}\textbf{Pairwise t-tests} & Notebook LM & 45/45 & 100.0 \\
    \rule{0pt}{10pt} & Grok3 & 35/45 & 77.8 \\
    \rule{0pt}{10pt}  & Sonnet4 & 24/45 & 53.3 \\
    \rule{0pt}{10pt}  & OpenAI-o3 & 10/45 & 22.2 \\
    \rule{0pt}{10pt} & ChatGPT-4.5 & 4/45 & 8.9 \\
    \rule{0pt}{10pt} & Gemini-2.5 & 1/45 & 2.2 \\
    \rule{0pt}{10pt}\textbf{Rank correlation tests} & Notebook LM & 21/21 & 100.0 \\
    \rule{0pt}{10pt} & Grok3 & 21/21 & 100.0 \\
    \rule{0pt}{10pt} & Sonnet4 & 1/21 & 4.8 \\
    \rule{0pt}{10pt} & OpenAI-o3 & 1/21 & 4.8 \\
    \rule{0pt}{10pt} & ChatGPT-4.5 & 0/21 & 0.0 \\
    \rule{0pt}{10pt} & Gemini-2.5 & 0/21 & 0.0 \\
    \bottomrule
    \end{tabularx}
    \caption{Significance testing results by multiple comparison correction}
\end{table}

\textit{Note: Cronbach's $\alpha$ coefficients and mean correlation coefficients are referenced in Appendix~\ref{appendix_A2}}

\subsubsection{Details of Correction Method}

\textbf{Benjamini-Hochberg Method Correction Procedure}:
\begin{enumerate}
    \item Sort all pairwise comparison p-values in ascending order
    \item Calculate adjusted q-value for each p-value: q(i) = p(i) $\times$ n / i
    \item Apply monotonicity constraint: q(i) = min(q(i), q(i+1))
    \item Judge q $<$ 0.05 as statistically significant
\end{enumerate}

\textbf{Test Settings}:
\begin{itemize}
    \item Type I error rate (false discovery rate): $\alpha$ = 0.05
    \item Used pairwise t-tests for paired samples
    \item Applied null correlation tests for rank correlations
    \item Total comparisons: 45 for pairwise t-tests, 21 for rank correlation tests
\end{itemize}

\section{APPENDIX - Detailed Language Analysis}
\label{appendix_E}

\subsection{Stopword Design}
\label{appendix_E1}
In the preprocessing for TF-IDF analysis, 50 stopwords were excluded based on the following selection criteria:
Selection criteria:

General function words: Words that have only grammatical functions without semantic features
Evaluation target words: Words that appear commonly across all evaluations and do not show differences between models (e.g., "narrative," "story," "ai," etc.)
Analysis noise words: Words appearing due to typographical errors or special notations (e.g., "fs," etc.)

\begin{itemize}
\item{\textbf{Function words}:}\\ the, a, an, in, on, of, to, with, at, by, for, from, is, are, was, were, be, being, been
\item{\textbf{Demonstratives and conjunctions}:}\\ that, this, these, those, and, but, or, if, then, so, as, because, while, although, though, yet, also, just
\item{\textbf{Adverbs and others}:}\\ it, its, fs, about, ai, narrative, story, through, slightly, somewhat, effectively, well, occasionally
\end{itemize}

\subsection{Model-specific Vocabulary Analysis Results}
\newcolumntype{Y}{>{\centering\arraybackslash}X}
\begin{table}[H]
    \centering
    \begin{tabularx}{\linewidth}{Y Y Y Y Y Y Y}
    \toprule
    \thead{\textbf{Rank}} & 
    \thead{\textbf{ChatGPT}\\\textbf{-4.5}} & 
    \thead{\textbf{Gemini}\\\textbf{-2.5}} & 
    \thead{\textbf{Notebook}\\\textbf{LM}} & 
    \thead{\textbf{OpenAI}\\\textbf{-o3}} & 
    \thead{\textbf{Grok3}} & 
    \thead{\textbf{Sonnet4}} \\
    \midrule
    \rule{0pt}{18pt} 1 & emotional & human & human & emotional & emotional & emotional \\
    \rule{0pt}{18pt} 2 & resonance & concept & consciousness & exposition & literary & human \\
    \rule{0pt}{18pt} 3 & cohesion & unsettling & resonant & prose & concept & philosophical \\
    \rule{0pt}{18pt} 4 & depth & humanity & psychological & originality & resonance & feels \\
    \rule{0pt}{18pt} 5 & originality & identity & identity & structural & original & exploration \\
    \rule{0pt}{18pt} 6 & exploration & original & mystery & character & prose & consciousness \\
    \rule{0pt}{18pt} 7 & literary & exploration & humanity & pacing & quality & identity \\
    \rule{0pt}{18pt} 8 & human & emotional & efficiency & overall & strong & character \\
    \rule{0pt}{18pt} 9 & strong & compelling & concept & concept & aesthetic & psychological \\
    \rule{0pt}{18pt}10 & existential & profound & original & piece & structure & compelling \\
    \bottomrule
    \end{tabularx}
    \caption{Characteristic evaluation vocabulary based on TF-IDF (Top 10 words)}
    \label{table_TF-IDF_model}
\end{table}

\textit{Note: Each vocabulary reflects evaluation approach characteristics}

\begin{itemize}
    \item ChatGPT-4.5: Constructive evaluation conscious of emotion-logic harmony
    \item Gemini-2.5: Evaluation exploring creativity and human existence boundaries
    \item Notebook LM: Philosophical and psychological depth exploration
    \item OpenAI-o3: Evaluation of stylistic beauty evoking emotion
    \item Grok3: Harmonious evaluation emphasizing literary resonance
    \item Sonnet4: Philosophy and emotion resonance as evaluation axes
\end{itemize}

\subsection{Cluster-specific Vocabulary Analysis Results (Supplementary Experiment)}
\textbf{Analysis Conditions}
\begin{itemize}
    \item Due to limited evaluation text samples, we did not standardize evaluation numbers but included all samples from each cluster for TF-IDF analysis
    \item Clusters based on 5 clusters obtained from K-means
    \item Notebook LM reduced to one session due to complete overlap
\end{itemize}

\newcolumntype{Y}{>{\centering\arraybackslash}X}
\begin{table}[H]
    \centering
    \begin{tabularx}{\linewidth}{Y Y Y Y Y Y}
    \toprule
    \thead{\textbf{Rank}\\\textbf{Cluster 0}} &
    \thead{\textbf{Cluster 1}\\\textbf{(70 cases)}} &
    \thead{\textbf{Cluster 2}\\\textbf{(130 cases)}} &
    \thead{\textbf{Cluster 3}\\\textbf{(30 cases)}} &
    \thead{\textbf{Cluster 4}\\\textbf{(70 cases)}} &
    \thead{\textbf{Cluster 5}\\\textbf{(60 cases)}} \\
    \midrule
    \rule{0pt}{18pt}1 & emotional & emotional & concept & emotional & human \\
    \rule{0pt}{18pt}2 & resonance & resonance & emotional & literary & emotional \\
    \rule{0pt}{18pt}3 & human & concept & identity & concept & exploration \\
    \rule{0pt}{18pt}4 & cohesion & human & original & resonance & feels \\
    \rule{0pt}{18pt}5 & exploration & originality & philosophical & original & philosophical \\
    \rule{0pt}{18pt}6 & depth & depth & psychological & prose & compelling \\
    \rule{0pt}{18pt}7 & originality & compelling & human & quality & identity \\
    \rule{0pt}{18pt}8 & concept & exploration & unsettling & strong & character \\
    \rule{0pt}{18pt}9 & existential & humanity & exploration & aesthetic & structure \\
    \rule{0pt}{18pt}10 & humanity & strong & abstract & structure & consciousness \\
    \bottomrule
    \end{tabularx}
    \caption{Cluster-specific TF-IDF characteristic Words and sample sizes (Top 10 words)}
    \label{table_TF-IDF_cluster}
\end{table}

\subsection{Interpretation of Vocabulary Analysis Results}

\subsubsection{Model-specific Evaluation Vocabulary Features}
Clear differences in evaluation frameworks are confirmed from each model's characteristic word analysis. ChatGPT-4.5's "cohesion" and "depth" indicate orientation toward structural evaluation, Gemini-2.5's "unsettling" and "humanity" show exploration of human existence, and Notebook LM's "consciousness" and "psychological" demonstrate psychological approaches.

\subsubsection{Inter-cluster Evaluation Profile Differences}
Cluster-specific vocabulary analysis conducted as supplementary experiment suggests the existence of evaluation profiles transcending models. Particularly in Cluster 1 (130 cases), "originality" and "humanity" appear as characteristic words, confirming the existence of evaluation profiles emphasizing innovation and humanity. In Cluster 4, "philosophical" and "consciousness" rank high, identifying evaluation profiles characterized by philosophical exploration.

\subsubsection{Model-Cluster Correspondence Relationships}
As a notable phenomenon, all Grok3 evaluations (7 sessions) are confirmed to belong completely to Cluster 4. Consequently, Grok3's characteristic words (top 10) from model-specific analysis completely match Cluster 4's characteristic words. This indicates that Grok3 forms a unique evaluation profile without mixing with other models.

\subsubsection{Hierarchical Feature Analysis of Evaluation Vocabulary}
Focusing on characteristic words ranked 4th and below in cluster-specific analysis reveals deeper characteristics as evaluators. For example, Cluster 2 features introspective and instability-suggesting vocabulary like "philosophical," "psychological," and "unsettling." Meanwhile, Cluster 4 shows more intuitive and sensory evaluation vocabulary like "feels," "compelling," and "identity" ranking high. These detailed vocabulary choices suggest the existence of latent evaluation characteristics and criteria that AI models possess toward literary works.

While statistical interpretation is limited by sample size constraints, the phenomenon of AI model evaluation profiles converging into evaluation profiles transcending model architecture is observed.

\section{Appendix - Research Limitations and Future Prospects}
\label{appendix_F}
The following tables present key experimental design trade-offs in our study, highlighting the methodological decisions that shaped our research. Each table contrasts alternative approaches and their respective advantages and limitations, providing a systematic overview of our research design considerations. In the tables below, $\circ$ indicates major advantages, while $\triangle$ highlights potential limitations or risks. Detailed numeric results and mitigation strategies for these limitations are discussed in the main text (\S~3.3, \S5.1).

\subsection{Research Design Limitations}
\textbf{1. Experimental Scope Constraints}
\begin{itemize}
    \item Limited sample of 6 models $\times$ 7 evaluations $\times$ 10 works and restriction to May 2025 model versions
    \item Statistical power limitations affecting generalizability and long-term evolution tracking
    \item For temporal proximity effects in evaluation sessions, see comparative analysis in Table~\ref{table_session_scheduling}
\end{itemize}
\textbf{2. Content-Related Limitations}
\begin{itemize}
    \item Genre limitations: External validity constraints due to Science fiction limitation
    \item Translation effects: Potential bias from single translation process
    \item Original language evaluation comparison not conducted
\end{itemize}
\textbf{3. Methodological Approach Limitations}
\begin{itemize}
    \item Observational vs. experimental design: Causal inference limitations
    \item Need for experimental manipulation verification in future studies
    \item Limited ability to isolate factors influencing evaluation behaviors
\end{itemize}
 
\subsection{Technical Implementation Constraints}
\label{appendix_F2}
\textbf{1. API Implementation Factors}
\begin{itemize}
    \item Commercial API parameters undisclosed (temperature settings, serving configurations)
    \item Platform dependency issues including potential caching and memory effects
    \item Technical variations in model serving that may influence evaluation consistency
    \item API behaviors may change without documentation, affecting research replicability 
\end{itemize}

\label{appendix_F3}
\subsection{Future Research Directions}
\textbf{1. Scale and Condition Variation Research}
\begin{itemize}
    \item Larger-scale model and work sets with increased evaluation frequency
    \item Evaluation stability across different time intervals (days, weeks, months)
    \item Comparison across multiple API implementation versions
    \item Universality verification of evaluation profiles across diverse genres
    \item Measurement of evaluation variation under different critical perspective prompts
\end{itemize}
\textbf{2. Mechanism Elucidation Research}
\begin{itemize}
    \item Identification of evaluation profile formation factors
    \item Systematic control experiments of temperature parameters
    \item Comparative analysis with human evaluators and alignment measurement
    \item Elucidation of cross-model evaluation profile formation mechanisms
    \item Investigation of relationships between evaluation profiles and training data/RLHF processes
\end{itemize}
\textbf{3. Applied Research}
\begin{itemize}
    \item Implementation verification in educational fields
    \item Practical application research in publishing industry
    \item AI ensemble criticism systems utilizing multiple evaluation profiles
    \item Development of evaluation frameworks incorporating diversity of critical perspectives
\end{itemize}
\textbf{4. Learning Data and Evaluation Target Relationship Studies}
\begin{itemize}
    \item Comparative research using both original and existing works
    \item Analysis of the relationship between "degree of prior learning" and evaluation characteristics
    \item Development of efficient methodological approaches for cross-genre research
\end{itemize}
\textbf{5. AI Value System Archaeology}
\begin{itemize}
    \item Systematic analysis of aesthetic and ethical value systems internalized by AI models
    \item Exploration of value formation through training corpus analysis
    \item Documentation of evaluation biases across different domains and content types
\end{itemize}
\subsection{Theoretical Development Possibilities}
\begin{table}[H]
    \newcolumntype{C}[1]{>{\centering\arraybackslash}m{#1}}
    \centering
    \footnotesize
    \begin{tabular}{|C{2.5cm}|C{3cm}|C{3cm}|C{3cm}|}
    \hline
    \rule{0pt}{1.2em}\textbf{Research Domain} & 
    \rule{0pt}{1.2em}\textbf{Current Status} & 
    \rule{0pt}{1.2em}\textbf{Future Development} & 
    \rule{0pt}{1.2em}\textbf{Expected Outcomes} \\
    \hline
    AI Criticism Studies Establishment &
    \begin{itemize}[leftmargin=*, itemsep=2pt, parsep=0pt]
        \item Basic concept presentation
        \item Foundational examples
        \item Preliminary reception
    \end{itemize} &
    \begin{itemize}[leftmargin=*, itemsep=2pt, parsep=0pt]
        \item Development as a new domain
        \item Integration with aesthetics
        \item Broader community engagement
    \end{itemize} &
    \begin{itemize}[leftmargin=*, itemsep=2pt, parsep=0pt]
        \item Interdisciplinary field creation
        \item New methodologies
        \item Curricular inclusion
    \end{itemize} \\
    \hline
    Evaluation Profile Theory &
    \begin{itemize}[leftmargin=*, itemsep=2pt, parsep=0pt]
        \item Five-type identification
        \item Textual variation mapping
        \item Initial validation
    \end{itemize} &
    \begin{itemize}[leftmargin=*, itemsep=2pt, parsep=0pt]
        \item Systematic axis classification
        \item Comparative calibration
        \item Reader alignment
    \end{itemize} &
    \begin{itemize}[leftmargin=*, itemsep=2pt, parsep=0pt]
        \item Predictive modeling
        \item Application to critique
        \item Insightful metric development
    \end{itemize} \\
    \hline
    AI Extension of Reader-Response Theory &
    \begin{itemize}[leftmargin=*, itemsep=2pt, parsep=0pt]
        \item Prototype application
        \item Response trace collection
        \item Reader-AI comparison
    \end{itemize} &
    \begin{itemize}[leftmargin=*, itemsep=2pt, parsep=0pt]
        \item Theoretical refinement
        \item Integrated frameworks
        \item Reflexive modeling
    \end{itemize} &
    \begin{itemize}[leftmargin=*, itemsep=2pt, parsep=0pt]
        \item New literary theory creation
        \item Empirical hermeneutics
        \item AI-assisted interpretation
    \end{itemize} \\
    \hline
    \end{tabular}
    \caption{Theoretical development framework for AI criticism studies}
\end{table}

\subsection{Experimental Design Trade-Off Analyses}
\label{appendix_F5}
\subsubsection{Continuous vs. Distributed Session Scheduling}
\label{appendix_F5.1}

\begin{table}[H]
    \newcolumntype{C}[1]{>{\centering\arraybackslash}m{#1}}
    \centering
    \footnotesize
    \begin{tabular}{|C{4cm}|C{4cm}|C{4cm}|}
    \hline
    \rule{0pt}{1.2em}\textbf{Evaluation aspect} & 
    \rule{0pt}{1.2em}\textbf{Continuous (1-day $\times$ 7)} & 
    \rule{0pt}{1.2em}\textbf{Distributed (days / weeks)} \\
    \hline
    Long-term drift suppression & $\circ$ Parameter / policy changes minimised & $\triangle$ Updates may occur between sessions \\
    \hline
    Practical burden & $\circ$ Data gathered quickly in one day & $\triangle$ Scheduling \& monitoring over days / weeks \\
    \hline
    Evaluation fatigue & $\triangle$ Risk of lexical rigidity after repetitions & $\circ$ Rest intervals mitigate fatigue \\
    \hline
    Platform transients & $\triangle$ Cache reuse / rate-limit windows may inflate variance & $\triangle$ Load-dependent restarts may affect answers \\
    \hline
    \end{tabular}
    \caption{Comparative analysis: temporal scheduling options}
    \label{table_session_scheduling}
\end{table}

\subsubsection{Original vs. Existing Works as Evaluation Material}
\label{appendix_F5.2}
\begin{table}[H]
    \newcolumntype{C}[1]{>{\centering\arraybackslash}m{#1}}
    \centering
    \footnotesize
    \begin{tabular}{|C{4cm}|C{4cm}|C{4cm}|}
    \hline
    \rule{0pt}{1.2em}\textbf{Evaluation aspect} & 
    \rule{0pt}{1.2em}\textbf{Original SF (this study)} & 
    \rule{0pt}{1.2em}\textbf{Existing published works} \\
    \hline
    Training overlap risk & $\circ$ Memory replay largely eliminated & $\triangle$ High overlap for popular works \\
    \hline
    Design control & $\circ$ Length / theme can be normalised & $\triangle$ Large variability requires extra normalisation \\
    \hline
    Ethics \& copyright & $\circ$ Rights fully under authors' control & $\triangle$ Permissions \& fair-use constraints \\
    \hline
    External quality anchors & $\triangle$ Few objective quality signals & $\circ$ Awards / criticism history available \\
    \hline
    Genre coverage & $\triangle$ High cost to craft original works across genres & $\circ$ Broad genre \& form coverage \\
    \hline
    \end{tabular}
    \caption{Comparative analysis: evaluation corpus selection}
    \label{table_eval_material}
\end{table}

\subsubsection{Default vs. Specialised Prompts}

\begin{table}[H]
    \newcolumntype{C}[1]{>{\centering\arraybackslash}m{#1}}
    \centering
    \footnotesize
    \begin{tabular}{|C{4cm}|C{4cm}|C{4cm}|}
    \hline
    \rule{0pt}{1.2em}\textbf{Evaluation aspect} & 
    \rule{0pt}{1.2em}\textbf{Default minimal prompt} & 
    \rule{0pt}{1.2em}\textbf{Specialised viewpoint prompt} \\
    \hline
    Implicit value-system exposure & $\circ$ Natural RLHF bias observable & $\triangle$ Obedience to directive may mask bias \\
    \hline
    Cross-model comparability & $\circ$ Single condition simplifies comparison & $\triangle$ Comparison confounded by prompt differences \\
    \hline
    RLHF policy detection & $\circ$ Autonomous safety / censorship visible & $\triangle$ Instructions may conflict with guardrails \\
    \hline
    Task adaptability & $\triangle$ Hard to probe specific critique angles & $\circ$ Allows targeted evaluation dimensions \\
    \hline
    Sensitive content handling & $\triangle$ Potential inconsistency in default mode & $\triangle$ Directive may trigger refusal / heavy redaction \\
    \hline
    \end{tabular}
    \caption{Comparative analysis: prompt design strategies}
    \label{table_prompts_strategies}
\end{table}

\subsubsection{Model-specific vs. Platform-related Effects}

\begin{table}[H]
    \newcolumntype{C}[1]{>{\centering\arraybackslash}m{#1}}
    \centering
    \footnotesize
    \begin{tabular}{|C{4cm}|C{4cm}|C{4cm}|}
    \hline
    \rule{0pt}{1.2em}\textbf{Evaluation aspect} & 
    \rule{0pt}{1.2em}\textbf{Model-inherent characteristics} & 
    \rule{0pt}{1.2em}\textbf{Platform implementation effects} \\
    \hline
    Evaluation consistency & $\circ$ Reflects true cognitive processing differences & $\triangle$ May be influenced by caching, memory effects \\
    \hline
    Internal validation & $\circ$ Characteristic vocabulary suggests deep patterns & $\triangle$ Technical API variations could cause inconsistency \\
    \hline
    Measurement reliability & $\circ$ Multiple evaluations provide robust sampling & $\triangle$ Session separation may not fully eliminate platform effects \\
    \hline
    Theoretical implications & $\circ$ Suggests internalized evaluation frameworks & $\triangle$ Some consistency patterns may be artifacts of implementation \\
    \hline
    Research replicability & $\triangle$ Model versions change over time & $\triangle$ API behaviors may change without documentation \\
    \hline
    \end{tabular}
    \caption{Comparative analysis: variance attribution factors}
    \label{table_variance_factors}
\end{table}

\bibliographystyle{plain}
\bibliography{references}

\end{document}